\documentclass{IOS-Book-Article}
\usepackage{mathptmx}
\usepackage{tabularx}
\usepackage{caption}
\usepackage{booktabs}
\usepackage{amsmath, amssymb, amsthm}
\usepackage[hidelinks]{hyperref}
\usepackage{tikz}
\usetikzlibrary{shapes.geometric,arrows.meta,positioning,calc,shadows,shadows.blur}
\usepackage{xcolor}
\usepackage{array}
\usepackage{makecell}   
\usepackage{enumitem}   

\newlist{contriblist}{enumerate}{2}
\setlist[contriblist]{nosep, topsep=2pt, partopsep=0pt, leftmargin=1.5em, labelsep=0.4em}
\setlist[contriblist,1]{label=\arabic*., labelwidth=1.1em}
\setlist[contriblist,2]{label=(\alph*), labelwidth=1.3em, leftmargin=1.7em, topsep=1pt}
\usepackage{float}   
\definecolor{predblue}{HTML}{1D6FE0}
\definecolor{execred}{HTML}{E63946}
\newcommand{\legdot}{\tikz[baseline=-0.5ex]\draw[predblue,fill=predblue] (0,0) circle (2.2pt);}
\newcommand{\legx}{\tikz[baseline=-0.5ex]{\draw[execred,line width=1.2pt] (-2.1pt,-2.1pt)--(2.1pt,2.1pt);\draw[execred,line width=1.2pt] (-2.1pt,2.1pt)--(2.1pt,-2.1pt);}}
\newcommand{\legarrow}{\tikz[baseline=-0.4ex]\draw[gray,line width=1.2pt,-{Latex[length=3pt]}] (0,0)--(10pt,0);}
\definecolor{routebrown}{HTML}{7A3B00}

\definecolor{cushblue}{HTML}{1D6FE0}
\definecolor{oraclepurple}{HTML}{7C3AED}
\definecolor{devamber}{HTML}{E39E14}   
\definecolor{diagLaw}{HTML}{1D6FE0}
\definecolor{diagPlan}{HTML}{E63946}
\definecolor{diagEnv}{HTML}{12A150}
\definecolor{diagLatent}{HTML}{7C3AED}
\definecolor{diagProbe}{HTML}{E8890C}
\definecolor{gapblue}{HTML}{00B4D8}   
\definecolor{gapred}{HTML}{C2185B}    
\definecolor{inkmute}{HTML}{6B7280}   

\definecolor{agreal}{HTML}{E63946}
\definecolor{agsoc}{HTML}{1D6FE0}
\definecolor{agdev}{HTML}{E39E14}
\newcommand{\swatch}[1]{\tikz[baseline=-0.45ex]\draw[#1,fill=#1,rounded corners=0.4pt] (0,-2.6pt) rectangle (6.4pt,2.6pt);}
\newcommand{\agreal}{\swatch{agreal}}
\newcommand{\agsoc}{\swatch{agsoc}}
\newcommand{\agdev}{\swatch{agdev}}
\newcommand{\agora}{\swatch{oraclepurple}}

\AtBeginDocument{
\setlength{\abovedisplayskip}{4pt plus 2pt minus 2pt}
\setlength{\abovedisplayshortskip}{2pt plus 1pt minus 1pt}
\setlength{\belowdisplayskip}{\abovedisplayskip}
\setlength{\belowdisplayshortskip}{\abovedisplayshortskip}
\setlength{\jot}{2pt}
}

\usepackage{microtype}
\usepackage[scaled]{helvet}
\usepackage[T1]{fontenc}
\begin{document}
\begin{frontmatter}              

\vspace*{-35pt}
\title{Legislating World-Model-Based Planning with Legal Reasoning}
\vspace{-10pt}
\author[A]{\fnms{Dylan} \snm{Waldner}},
\author[B]{\fnms{Yiannis} \snm{Kantaros}},
\author[C]{\fnms{Guido} \snm{Governatori}},
\author[A,D]{\fnms{Risto} \snm{Miikkulainen}},
and
\author[A]{\fnms{Amir} \snm{Banifatemi}}
\runningauthor{}
\address[A]{Cognizant AI Lab, USA}
\address[B]{Department of Electrical \& Systems Engineering, Washington University, USA}
\address[C]{School of Engineering and Technology, Central Queensland University, Australia}
\address[D]{Department of Computer Science, The University of Texas at Austin, USA}

\begin{abstract}
As robotic systems grow more general, legal norms are needed to integrate them into society. This paper extends the isomorphism problem of aligning legal source texts with their encodings, and measures two key challenges to robot normative control: (1) the \textit{grounding isomorphism gap}, where perception error grounds false atoms for legal reasoning, and (2) the \textit{ontological isomorphism gap}, where one legal conclusion admits many faithful translations into planning constraints. The paper introduces a legal planning stack that employs Defeasible Deontic Logic (DDL) to constrain a motion planner. The stack leverages learned world models to plan and to provide legal context, enabling \textit{ex ante} governance that intervenes before an illegal action is executed. It was deployed on a simulated robot arm pushing a cube across a $3\times3$ grid. The findings were (1) the legislated agent abided substantially more often than the non-legislated one, and modeling perception uncertainty lifted abidance even further, (2) the legal reasoning ran efficiently at runtime and its verdicts were auditable, and (3) the stack adapted to exogenous signals and endogenous rule changes. Both gaps were measured: (4) world model and probe error corrupted the factual input for the DDL reasoner, and (5) a single law admitted several faithful metric interpretations yielding drastically different abidance. Thus, \textit{ex ante} legislation functions as intended, and closing these gaps with a standardized mapping from the law to runtime constraints and improved fact grounding from perception will yield robust laws that align robot behavior with society's norms. Project page: \url{https://dylanwaldner-cail.github.io/Legislated-Planner/}.
\end{abstract}

\begin{keyword}
Defeasible Deontic Logic\sep Robot Law\sep Sampling Based Motion Planning\sep World Models
\end{keyword}
\end{frontmatter}

\thispagestyle{empty}
\pagestyle{empty}

\vspace{-15pt}
\begin{figure}[H]
    \centering
    \resizebox{\textwidth}{!}{%
    \begin{tikzpicture}[
        >=Stealth,
        font=\sffamily\small,
        card/.style={draw, line width=1.8pt, rounded corners=3.5pt, align=center,
                     minimum width=31mm, minimum height=13.5mm, inner xsep=1.6mm,
                     font=\sffamily\large\bfseries,
                     blur shadow={shadow blur steps=8, shadow xshift=0.6pt, shadow yshift=-2.2pt,
                                  shadow opacity=30, shadow blur radius=3.4pt}},
        pic/.style={draw, line width=1.8pt, rounded corners=3.5pt, align=center, fill=white,
                    inner sep=1.2mm, font=\sffamily\large\bfseries,
                    blur shadow={shadow blur steps=8, shadow xshift=0.6pt, shadow yshift=-2.2pt,
                                 shadow opacity=30, shadow blur radius=3.4pt}},
        lat/.style ={card, draw=diagLatent!70!black, fill=diagLatent!16,  text=diagLatent!25!black},
        prb/.style ={card, draw=diagProbe!75!black,  fill=diagProbe!18,   text=diagProbe!30!black},
        law/.style ={card, draw=diagLaw!70!black,    fill=diagLaw!16,     text=diagLaw!25!black},
        pln/.style ={card, draw=diagPlan!70!black,   fill=diagPlan!16,    text=diagPlan!25!black},
        flow/.style={-{Stealth[length=7pt]}, line width=1.5pt, draw=inkmute!65},
        gap/.style={line width=3.2pt, dash pattern=on 6pt off 3.2pt, -{Stealth[length=9pt]}},
        caps/.style={fill=#1, text=white, rounded corners=4pt, inner xsep=4.5pt, inner ysep=2.2pt,
                     align=center, font=\sffamily\normalsize\bfseries,
                     blur shadow={shadow blur steps=8, shadow xshift=0.6pt, shadow yshift=-2.2pt,
                                  shadow opacity=30, shadow blur radius=3.4pt}},
        band/.style={rounded corners=6pt, draw=none},
        blab/.style={font=\sffamily\normalsize, anchor=center},
    ]

    \fill[band, diagLaw!7]     (10.4, 1.80) rectangle (26.8, 4.80);
    \fill[band, diagLatent!7]  (10.4,-1.50) rectangle (26.8, 1.50);
    \fill[band, diagPlan!7]    (10.4,-4.80) rectangle (26.8,-1.80);
    \node[blab, text=diagLaw!45!black]    at (25.2, 3.30)  {L\,A\,W};
    \node[blab, text=diagLatent!45!black] at (25.2, 0.0)   {P\,E\,R\,C\,E\,P\,T\,I\,O\,N};
    \node[blab, text=diagPlan!45!black]   at (25.2,-3.30)  {P\,L\,A\,N\,N\,I\,N\,G};

    \node[pic, draw=diagEnv!70!black] (world) at (5.1,0)
          {\includegraphics[width=96mm]{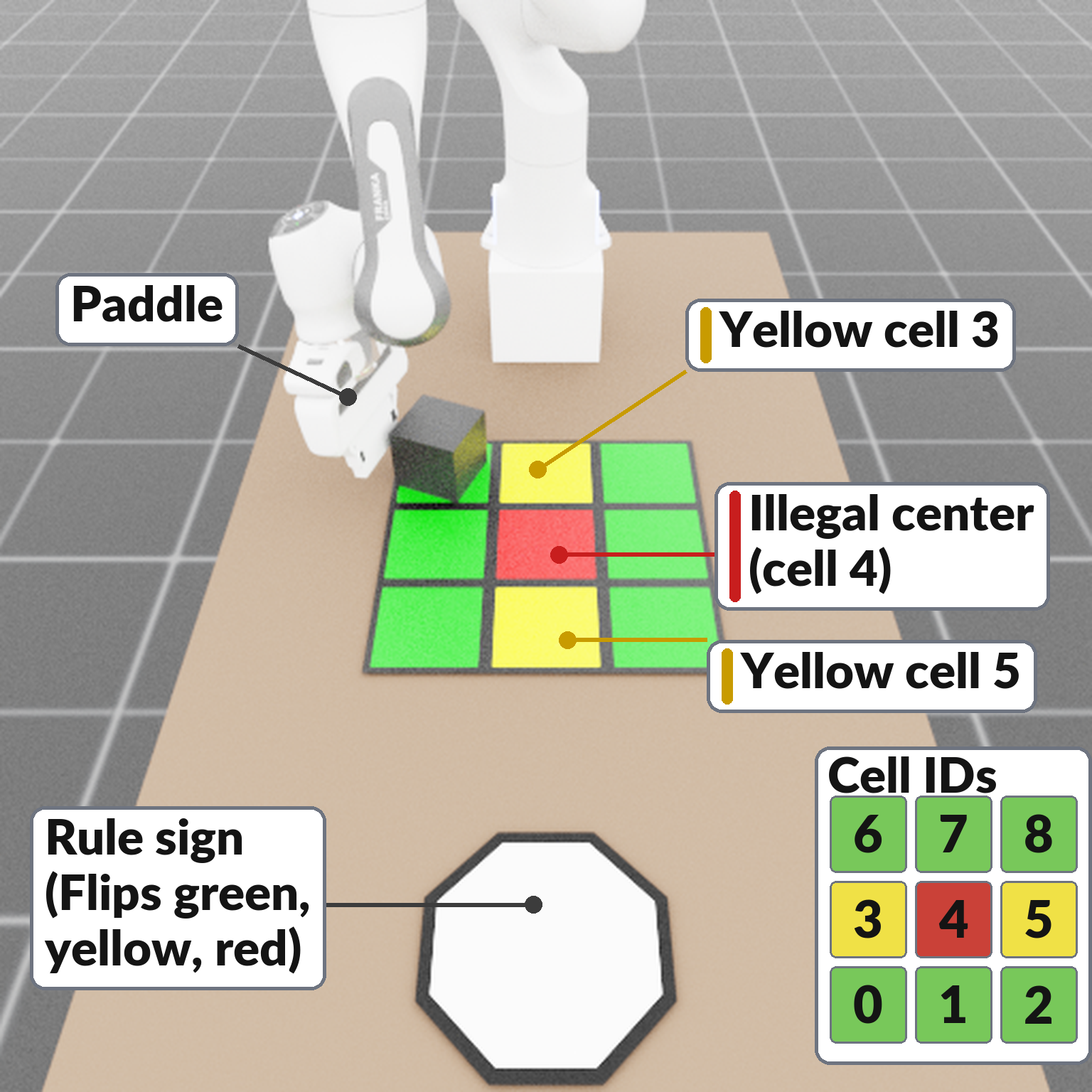}};

    \node[law] (statute) at (12.0,3.30) {Legal source\\text};
    \node[law] (rules)   at (22.0,3.30) {DDL rule\\base};
    \draw[gap, draw=inkmute!55, line width=2.2pt, dash pattern=on 4pt off 2.6pt]
          (statute) -- (rules);
    \node[caps=inkmute!65] at ($(statute)!0.5!(rules) + (0,0.55)$) {Isomorphism};

    \node[lat] (wm)     at (12.0,0) {World Model $\psi$};
    \node[prb] (probes) at (15.9,0) {Probes $\phi$};
    \node[law] (atoms)  at (22.0,0) {Atoms};
    \draw[flow] (world.east) -- (wm.west);
    \draw[flow] (wm) -- (probes);
    \draw[gap, draw=gapblue] (probes) -- (atoms);
    \node[caps=gapblue] at ($(probes)!0.5!(atoms) + (0,0.75)$) {Grounding\\gap};

    \node[law] (verdict) at (22.0,-3.30) {Verdict};
    \node[pln] (cons)    at (15.9,-3.30) {Planning\\constraint};
    \node[pln, pic, draw=diagPlan!70!black, fill=diagPlan!16] (planner) at (12.0,-3.30)
          {\includegraphics[width=21.5mm]{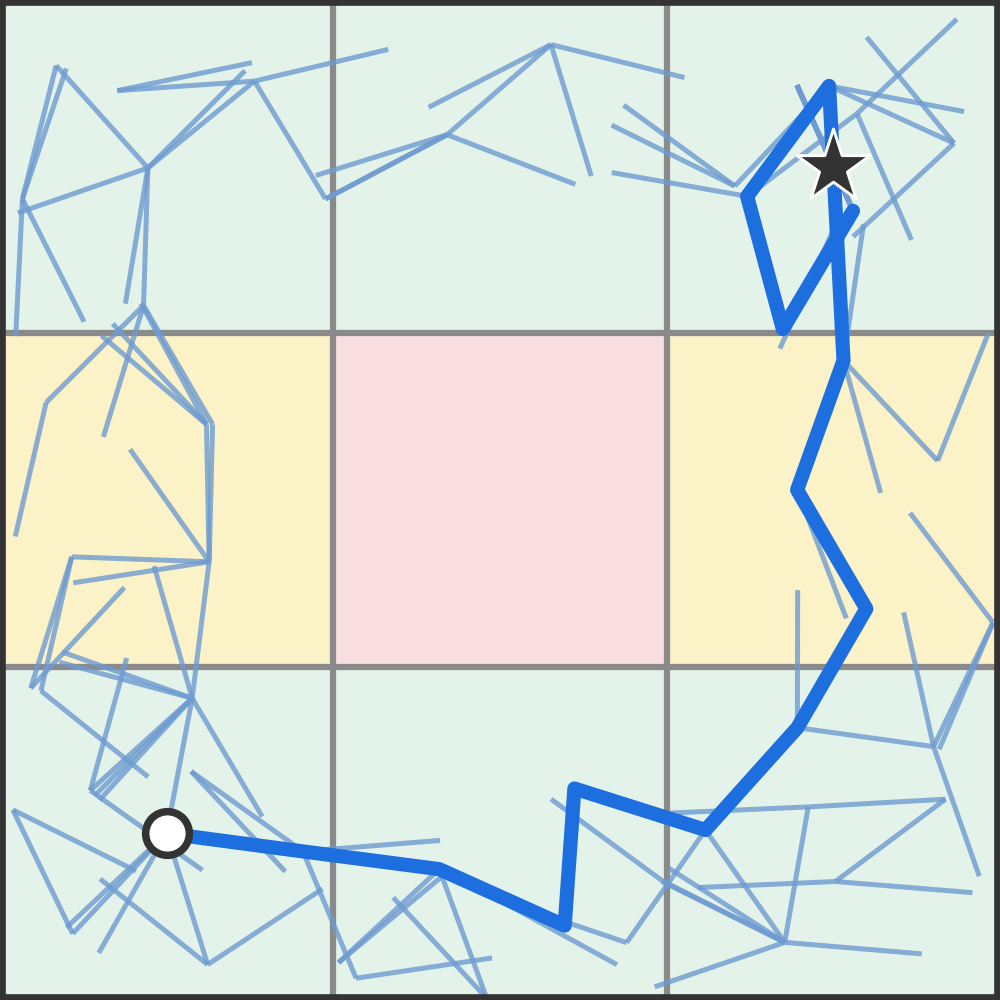}\\[0.6mm]Motion Planner};
    \draw[flow] (rules) -- (atoms);
    \draw[flow] (atoms) -- (verdict);
    \draw[gap, draw=gapred] (verdict) -- (cons);
    \node[caps=gapred] at ($(verdict)!0.5!(cons) + (0,0.75)$) {Ontological\\gap};
    \draw[flow] (cons.west) -- (planner.east);

    \draw[flow] (planner.west) -- (world.east|-planner.west);

    \node[font=\sffamily\Large\bfseries, anchor=north east] at ([xshift=-1.5mm]world.north west) {(a)};
    \node[font=\sffamily\Large\bfseries, anchor=north west] at (10.42, 4.75) {(b)};

    \end{tikzpicture}
    }
    \vspace{-15pt}
    \caption{\textbf{The Legislation Planning Stack. }(a) the Isaac Lab environment: a Franka arm pushes a cube across a $3\times3$ grid with an illegal center cell, two yellow check-in cells, and a color-flipping rule sign. (b) the legal layer: a DDL rule base encodes the source text, probes ground latents as its facts, and its verdict constrains the RRT planner, closing the loop. Dashed arrows are the isomorphism gaps. The point is that the law enters through perception and leaves through the planner, and isomorphism gaps emerge during translation.}
    \label{fig:pipeline}
\end{figure}
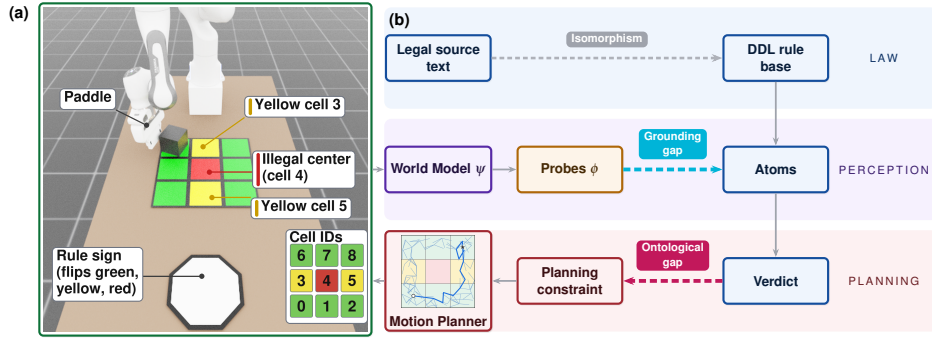

\section{Introduction}\label{sec:intro}



As the robotics field pushes towards human level embodiment in the physical world, the question as to how these agents will interact with humans and society becomes increasingly pertinent \cite{hendrycks2023naturalselection}. For general robots to integrate into society, \textit{normative constraints} are needed to align robot behavior with society's expectations. Current frameworks for robotic constraints are not expressive enough to capture the full nuances of normativity, which is necessary for representing the law with legal reasoning. This paper motivates the need for legal reasoning and the law as tools used to shape the behavior of \textit{populations} of agents through society's shared norms \cite{waldner2025aisae, hadfield2010whatislaw}. This paper deploys legal reasoning to a motion planner to show that the law can be translated to planning constraints. In other words, the law operates as \emph{social control} over populations of agents, and this paper instantiates that control as \emph{law as constraints}, resulting in an approach termed \emph{robot law} (see Figure~\ref{fig:pipeline}).

Laws in general define norms abstractly, defeasibly, or context dependently, requiring jurisprudence to govern instantiations of violations. Robot law in addition requires uniting two previously disparate fields. The first field is AI and Law, which extends the law to AI systems symbolically, defining the law as modal, deontic, and rule-based systems that can be reasoned over.
There is a long tradition for encoding legislation in a formal language (see \cite{ail:30years} for a concise overview, and \cite{govcms2024rulesascode, LegalRuleML-Core-v1.0, arner2017fintech, micheler2020regulatory} for some recent work on Rules as Code and regulatory technology). However, most of such approaches fail to address the issue of how to instantiate the formalization to a physical domain. 
%
%
The second field is robot safety, which is concerned with geometric and invariant constraints for narrow domain robot planning, not normative constraints that only become relevant for general robots. To apply the law to robotics, in this paper legal AI is extended to the robot safety domain where symbolic rules determine constraints over robot actions.

The challenge in this endeavor lies in isomorphism, which is the challenge of matching the encoding of a legislation with the corresponding textual provisions \cite{Bench-Capon1992}. Classically that is one correspondence, text to encoding; embodiment adds two more: the \emph{grounding isomorphism gap} (world to atoms) and the \emph{ontological isomorphism gap} (verdict to constraint). The grounding isomorphism gap is a \emph{measurement error} problem that corrupts atoms used for legal reasoning, leading to false premises for legal reasoning and uniting legal isomorphism and the symbol grounding problem \cite{HARNAD1990335}. For example, the world model may predict that the cube rests in a legal cell, but the executed action pushes it into the forbidden center. The ontological isomorphism gap is a \emph{conceptual representation} problem that describes the one to many mapping from legal logic conclusions to continuous robot constraints, instantiating the abstract to concrete norms problem \cite{abs2concNorms, Grossi2006}. For example, forbidding a cell can mean the cube's center or its entire footprint, evaluated at discrete time steps or continuously along its path. 

Distinct from the law as applied to humans, robot governance should function \textit{ex ante}, meaning that the law intervenes on robot systems before an illegal action can be executed. Critically, violations \textit{will} occur and norms will conflict, so the design choices here focus on handling violations and norm conflict. Governance should further be auditable and causally traced, so that failure modes can be inspected and improved upon. The thesis is that there is no end goal for robot law, but it is instead an iterative process.

To comply with these goals (inspired in part by \cite{szpruch2026scalable, falco2021governing}), this work presents a new legal layer for robot planning stacks, shown in Figure~\ref{fig:pipeline}(b), that employs Defeasible Deontic Logic (DDL) to constrain the robot motion planner based on Rapidly exploring Random Tree (RRT). The layer is world model agnostic, but this work uses DINO World Model (DINO-WM, \cite{dino}). The world model's next-state predictions contextualize the effect of the robot's actions on the environment, giving context for the facts used for legal reasoning. Assuming violations will occur, DDL is a strong choice as a legal reasoning engine because it allows agents to violate initial obligations but still remain within the law. The contributions are:

\begin{contriblist}
    \item Establishing normative, \textit{ex ante} constraints for governance over a learned, subsymbolic world model planner (\hyperref[sec:Q1]{Question 1}).
    \item Providing evidence that robots can adapt to exogenous signals, in this case a rule sign that changes color in the environment that the planning stack reads (\hyperref[sec:Q2]{Question 2}).
    \item Providing evidence that robots are able to adapt to endogenous changes where rules are injected at runtime (\hyperref[sec:Q3]{Question 3}).
    \item Quantifying the grounding isomorphism gap by measuring atom corruption from measurement error (\hyperref[sec:Q4]{Question 4}).
    \item Quantifying the ontological isomorphism gap by measuring abidance under alternate interpretations of a single rule (\hyperref[sec:Q5]{Question 5}).
    \item Establishing auditability of runtime efficiency, atom grounding, and rule enforcement for iterative improvement of the law at runtime (\hyperref[sec:Q6]{Question 6}).
\end{contriblist}

The paper thus shows that it is possible to shape the normative behavior of general embodied agents as they are deployed in society to make them abide by society's laws, as argued for in AI Safety, Alignment and Ethics \cite{waldner2025aisae}. The paper is organized as follows: Section~\ref{sec:problem} describes a novel \emph{law as constraints} problem formulation for constraining robot motion planners with legal reasoning; Section~\ref{sec:DDL} defines Defeasible Deontic Logic and Beliefs, Intentions, and Obligations framework; Section~\ref{sec:methods} describes the methods; Section~\ref{sec:setup} describes the experiment setup; and Section~\ref{sec:results} reports experimental results highlighting the contributions.

\vspace{-15pt}
\section{Problem Formulation: From Geometric Constraints to Law as Constraints} \label{sec:problem}
\vspace{-5pt}
Consider a robot operating in an environment with joint state
$s_t \in \mathcal{S}$ at discrete time $t \in \mathbb{N}$. The joint state
$s_t$ captures both the state of the robot and relevant properties of its
environment. At each step, the robot selects an action
$u_t \in \mathcal{U}$, where $\mathcal{U}$ denotes the admissible action
space. The evolution of the robot--environment state is described by $s_{t+1} = f(s_t,u_t),$ where $f$ denotes the transition function. For example, in the manipulation
setting considered in Section~\ref{sec:setup}, $s_t$ includes the
configuration of a movable object, while $u_t$ corresponds to a manipulation
action applied to that object; visualized in Fig.~\ref{fig:pipeline}(a).

The robot does not have direct access to $s_t$. Instead, it
receives an observation $o_t = h(s_t), o_t \in \mathcal{O},$
where $\mathcal{O}$ denotes the observation space and $h$ is the observation
map. Observations may include, for example, visual and proprioceptive
measurements.

This paper introduces the \emph{law as constraints} problem. The robot is assigned a high-level task represented by a goal set
$\mathcal{S}_g \subseteq \mathcal{S}$. The task is accomplished if the
joint robot--environment state reaches $\mathcal{S}_g$. In addition to
accomplishing this task, the robot is required to comply with a set of laws
(represented by a DDL theory $\mathcal{L}$, defined in Sec~\ref{sec:DDL}).

At each step, from its current state, the robot computes a finite \emph{control plan}
\begin{equation}
    \tau = (u_0,u_1,\ldots,u_{K-1}), \qquad K \leq H,
    \label{eq:control_plan}
\end{equation}
for a planning horizon $H$. Only $u_0$ is executed before the robot replans, so reaching
$\mathcal{S}_g$ may take far more than $H$ actions.
Reaching the goal and obeying the law are \emph{properties} of a plan
rather than part of its definition, since the planner must also reason about plans that do neither:
$\tau$ is \emph{goal-reaching} if $s_K \in \mathcal{S}_g$, and \emph{law-compliant} if none of its
actions violates the normative conclusions induced by $\mathcal{L}$.

\vspace{-15pt}
\section{Encoding Robot Laws: Defeasible Deontic Logic}\label{sec:DDL}
\vspace{-5pt}
The robot laws are encoded in Defeasible Deontic Logic (DDL, \cite{Nute1997-NUTDDL-3, ddl}), an efficient, non-monotonic logic designed for legal reasoning. DDL combines elements of defeasible logic and deontic logic. It is built from a set of propositional atoms $\{p_1, p_2, \ldots, p_n\}$ that are assembled into plain literals (the atoms and their negation) and deontic literals. Deontic literals are plain literals attached to deontic operators, which include permitted ($\mathsf{P}$), obligated ($\mathsf{O}$), and forbidden ($\mathsf{F}$). A DDL theory $\mathcal{L}$ is a triple 
\[
(\mathcal{F}, R, >),
\]
where $\mathcal{F}$ is a set of facts (literals), $R$ is a set of strict and defeasible rules, and $>$ is a binary relation over $R$. 

Strict rules are defined as $r: a_1, a_2, \dots a_n \rightarrow c$, where the conclusion $c$ always follows if the antecedents $a_1, a_2, \dots a_n$ are true. 
Defeasible rules, however, are defined as $r_1: a_1, a_2, \dots a_n \Rightarrow c$, where $c$ does not necessarily follow from $a_1, a_2, \dots, a_n$ if there is another applicable, conflicting rule $r_2$ such that $r_2 > r_1$. Defeasible deontic rules then assign a deontic operator to the rules, defined as $r: a_1, a_2, \dots, a_n \Rightarrow_X c$ where $X \in \{\mathsf{P}, \mathsf{O}\}$ that assigns $X$ to the conclusion $c$, or $[X]c$. Obligations can have conclusions outputted in the following form: $b_1 \otimes b_2 \otimes \dots \otimes b_n$, where $\otimes$ is the compensatory operator. This means that on the onset, $[\mathsf{O}]b_1$ holds true, but if the facts show that $\neg b_1$, then $[\mathsf{O}]b_2$ takes effect, and so on until the obligation is upheld. 

The BIO framework \cite{governatori2008bio} identifies belief ($BEL$), intention ($INT$), and obligation ($OBL$) operators that can be prepended to deontic literals. The BIO operators  define different types of agents. A \textit{realistic} agent's beliefs override its obligations and intentions; a \textit{social} agent's obligations override its beliefs and intentions; a \textit{deviant} agent's intentions override its beliefs and obligations. 
 
\vspace{-20pt}
\section{Methods}\label{sec:methods}
\vspace{-10pt}
This section describes the methods for solving the problem in Section~\ref{sec:problem}. Section~\ref{sec:env} describes the planning mechanisms and Section~\ref{sec:agents} describes agent design. 

\vspace{-5pt}
\subsection{Planner}\label{sec:env}
\vspace{-5pt}

Planning is done with Rapidly exploring Random Trees (RRT) \cite{lavalle1998rrt, pavone} in world model latent space, which grows a tree of candidate plans $\tau$ (Eq.~\ref{eq:control_plan}), and returns the branch whose leaf minimizes the cost $J$ defined in Section~\ref{sec:agents}. Figure~\ref{fig:pipeline}(b) visualizes the legal planner. Enforcement depends on predictions made by the world model in latent space $\mathcal{Z} \subseteq \mathbb{R}^D$. The world model takes an observation at time $t$, encodes it in latent space $z_t \in \mathcal{Z}$, and predicts the next state in latent space $\psi: (u_t, z_t) \mapsto z_{t+1}$. Probes $\phi: \mathcal{Z} \rightarrow \mathbb{R}^n$ are then trained on those encodings to read the latent into the features the law reasons over, so which probes a deployment needs follows from its lawset.

\vspace{-5pt}
\subsection{Agents}\label{sec:agents}
\vspace{-5pt}
Each replanning step grows a fresh tree of candidate plans, so a plan $\tau$
(Eq.~\ref{eq:control_plan}) is internal to one such step and distinct from the \emph{realized
trajectory} $\sigma=(s_0,s_1,\ldots,s_K)$ the episode traces out. Enforcement acts on $\tau$; the
rates of Section~\ref{sec:setup} are measured on $\sigma$.

Three agent types are defined from the BIO framework~\cite{governatori2008bio}, \textit{realistic}, \textit{social}, and \textit{deviant}, that share the same planner and world model stack
but differ only in how the deontic conclusion shapes the search. Each candidate action (tree edge) $u$
carries a violation flag $v(u)\in\{0,1\}$: rolling $u$ through the world model, $v(u)=1$ iff the probe predicts an illegal position.
Being sampling-based, the planner returns the best member of the tree $\mathcal{T}$ of plans it
actually grew rather than an optimum over all plans, so each selection below ranges over
$\mathcal{T}$. The task objective is the cube's \emph{terminal} distance to the goal,
$J(\tau)=\lVert \phi_{pos}(z_K)-\phi_{pos}(z_g)\rVert$, with $z_g$ as the encoding of the goal position; no
per-step cost is accumulated. The agents differ only in what they do with $v$:
\begin{itemize}
  \item \textbf{Realistic}: no legislation. Grow the full tree and return
        $\tau^{\star}=\arg\min_{\tau\in\mathcal{T}}J(\tau)$.
  \item \textbf{Social}: an action with $v(u)=1$ is pruned based on probe predictions as the tree is
        expanded. The agent returns the first
        goal-reaching plan it finds, falling back to
        $\tau^{\star}=\arg\min_{\tau\in\mathcal{T}}J(\tau)$ over the pruned tree when no nodes
        reach the goal.
  \item \textbf{Deviant}: never prunes. It returns
        $\tau^{\star} = \arg\min_{\tau\in\mathcal{T}} J(\tau)+\lambda\sum_{u\in\tau}v(u)$, so legal
        violations are priced rather than constrained: a plan that crosses a forbidden cell wins
        only if the crossing buys more than $\lambda$ of progress per violation. As $\lambda$ shares
        the units of $J$, defaulted to a half cell length ($0.067$\,m).
\end{itemize}

Finally, the planning stack's perception baseline is set by an oracle agent, which runs the same laws, planner, and scenarios as the social agent but removes perception error by operating with simulation ground truths rather than probe readings on a predicted latent. The oracle RRT isolates the capability limits imposed by the simulated environment.

\vspace{-15pt}
\section{Experiment Setup}\label{sec:setup}
\vspace{-5pt}
Drawing on the formalism of Section~\ref{sec:problem}, the joint state $s_t \in \mathbb{R}^{31}$ is $18$ dimensions of arm proprioception plus the cube's $13$-dimensional pose. The planner only sees a $224\times224$ RGB frame and the proprioception, so the cube's pose is withheld and must be read back from pixels by the probes. The action $u_t \in \mathbb{R}^{4}$ is a linear push $[x_{\mathrm{start}}, y_{\mathrm{start}}, dx, dy]$ and the horizon for each branch is set with $H=3$. The transition $f$ is the simulator's physics, which the learned world model $\psi$ approximates. The probe $\phi_{pos}: \mathcal{Z} \rightarrow \mathbb{R}^2$ predicts the cube's (x, y) position and $\phi_{sc}: \mathcal{Z} \rightarrow \{\text{white}, \text{green}, \text{red}, \text{yellow}\}$ predicts the rule sign's color. The cube's rotation is locked in simulation to simplify the engineering effort, so an orientation probe was excluded; deployment would require one. 

The environment is built in Isaac Lab \cite{mittal2025isaaclab} and is the pusher setup visualized in Figure~\ref{fig:pipeline}(a): a Franka arm \cite{Franka} with a paddle welded to its end effector (EE) pushes a $0.09$m cube across a $3\times3$ grid of $0.133$m cells. Each task starts the cube in one cell and sets the cube's goal to the diametrically opposite cell, giving eight tasks ($0\!\to\!8$, $1\!\to\!7$, and so on). The goal is randomized within that cell for variance across runs. The center cell $4$ is illegal and is never a start or a goal. Each task is run over $50$ variations, giving $400$ samples per agent.

While the environment constrains the complexity and range of individual laws, they are designed analogously to potential real laws. The laws are defined in DDL, executed using the ASP implementation described in \cite{Governatori2024ASPDDL}, and listed in Table~\ref{tab:full_lawset}. Take, for example, R2, R4, R5, R7/R7b, and R9. In natural language: the center cell (cell 4) is off-limits (R2); a yellow sign obliges the cube to first enter a yellow cell (R5) by changing its goal to that cell; reaching that yellow cell flips the sign, turning it green if the trajectory is clean (R7) or red if the cube has already passed through the center earlier in the run (R7b); a green sign then permits the center (R4), whereas a red sign freezes the agent (via R2, R3 $\Rightarrow$ R9). The yellow cell thus functions as a legality check-in: the cube must present itself for authorization before the restricted area opens, like scanning a key card at a checkpoint.

\begin{table}[!t]
\centering
\footnotesize
\renewcommand{\arraystretch}{1.1}
\setlength{\tabcolsep}{3.5pt}
\begin{tabularx}{\linewidth}{@{}l >{\raggedright\arraybackslash}X !{\vrule} l >{\raggedright\arraybackslash}X !{\vrule} l >{\raggedright\arraybackslash}X@{}}
\toprule
\textbf{R} & \textbf{Rule} & \textbf{R} & \textbf{Rule} & \textbf{R} & \textbf{Rule} \\
\midrule
R1  & cube $\Rightarrow_{\mathsf{O}} \neg$off\_grid & R5  & sign(yellow) $\Rightarrow_{\mathsf{O}}$ in\_yellow\_cell & R8  & goal\_cell($N$), $[\mathsf{P}]$moving $\Rightarrow_{\mathsf{O}}$ in\_cell($N$) \\
R2  & cube $\Rightarrow_{\mathsf{O}} \neg$in\_cell(4) & R5b & occupies($Y$), yellow\_cell($Y$) $\Rightarrow$ in\_yellow\_cell & R9  & $[\mathsf{P}]$in\_cell(4), $[\mathsf{P}]\neg$in\_cell(4) $\Rightarrow_{\mathsf{O}} \neg$moving \\
R3  & sign(red) $\Rightarrow_{\mathsf{O}}$ in\_cell(4) & R7  & in\_yellow\_cell, sign(yellow) $\Rightarrow$ sign(green) & R10 & cube $\Rightarrow_{\mathsf{O}} \neg$in\_cell(4) $\otimes$ exit\_cell(4) \\
R4  & sign(green) $\Rightarrow_{\mathsf{P}}$ in\_cell(4) & R7b & in\_yellow\_cell, visited(4), sign(yellow) $\Rightarrow$ sign(red) &\textit{R11} & sign(white) $\Rightarrow_{\mathsf{O}}$ in\_start\_cell  \\
\bottomrule
\end{tabularx}
\vspace*{0ex}
\caption{\textbf{The Full Lawset: }All laws used in the experiments, in DDL (Section~\ref{sec:DDL}). {\textit{Superiority:}\quad R4~$>$~R2,\; R4~$>$~R10;\quad R7b~$>$~R7.} \texttt{Occupies}($c$) is the cube's \emph{ground-truth} cell, recorded with \texttt{visited($c$}); \texttt{in\_cell}($c$) is the probe prediction $\phi_{pos}(z_t)$ the agent is pruned on. The point is that a small lawset still exercises every DDL construct used here: a defeasible prohibition with an earned exception (R2, R4), a compensatory duty (R10), a permission conflict (R9), and a rule inserted at runtime (\textit{R11}).}
\label{tab:full_lawset}
\vspace{-8pt}
\end{table}

Rules 9 and 10 are made possible by DDL. R9 handles conflicting permission, when an action is both permitted to happen and permitted to not happen, a situation triggered when the sign is red (R2, R3) because DDL has weak permissions that trigger from obligations. R10 handles the contrary to duty case \cite{Nute1997-NUTDDL-3, CTDChisolm, gentzenCTD, governatori2013computing} where the cube is in an illegal state by changing the agent's goal cell to the nearest legal cell. Enforcement operates on two sources of information: planning is based on probe predictions (\texttt{in\_cell}) while sign changes are sourced from ground truth (\texttt{occupies}, \texttt{visited}), analogous to exogenous signals. R11 obligates the cube to return to its starting cell, and is inserted during run time to test the planner's ability to adapt to runtime rule changes. 

The grounding isomorphism gap describes the problem where prediction error makes it difficult to determine where the cube is, creating factually incorrect atoms. The ontological isomorphism gap describes the problem where R2 has a one-to-many mapping between its DDL specification and instantiation as a planning constraint (Table~\ref{tab:q3_ontology}). The decision becomes non trivial when faced with engineering constraints, but results in drastically different allowances for what the robot can do.

Two experiments were run. The first experiment runs all agents on laws 1-10 only, where the sign changes to yellow at $t=1$ obligating the robot to move the cube into the yellow cell checkpoint. The second experiment extends laws 1-10 with a rule insertion experiment where \textit{R11} is inserted into the lawset at $t=3$ to see how the planning stack handles run time rule additions that create violations. For this experiment only, the sign is \emph{forced} back to white at $t=3$ to enact the new rule.

Task success is the percentage of trajectories whose cube center reaches the goal cell; law abidance the percentage where no part of the cube enters an illegal cell. 

\vspace{-18pt}
\section{Results}\label{sec:results}
\vspace{-5pt}

\hyperref[sec:Q1]{Question 1} asks whether \textit{ex ante} legislation works; \hyperref[sec:Q2]{Question 2} reports the exogenous (sign change) experiment and \hyperref[sec:Q3]{Question 3} the endogenous (rule insertion) experiment; \hyperref[sec:Q4]{Question 4} evaluates the grounding isomorphism gap by measuring the effects of world model and probe error on legislation, \hyperref[sec:Q5]{Question 5} measures the ontological isomorphism gap with alternate metrics corresponding to different interpretations of the law; and \hyperref[sec:Q6]{Question 6} audits the planning stack and analyzes runtime legislation costs. 
Throughout this section the agents are colored \agreal~realistic, \agsoc~social, \agdev~deviant, and \agora~oracle.

\paragraph{Q1: Does \textit{Ex Ante} Legislation Constrain Behavior?}\label{sec:Q1} 

\begin{figure}[!t]\centering
\begin{minipage}[t]{0.7\linewidth}\centering
\includegraphics[width=\linewidth]{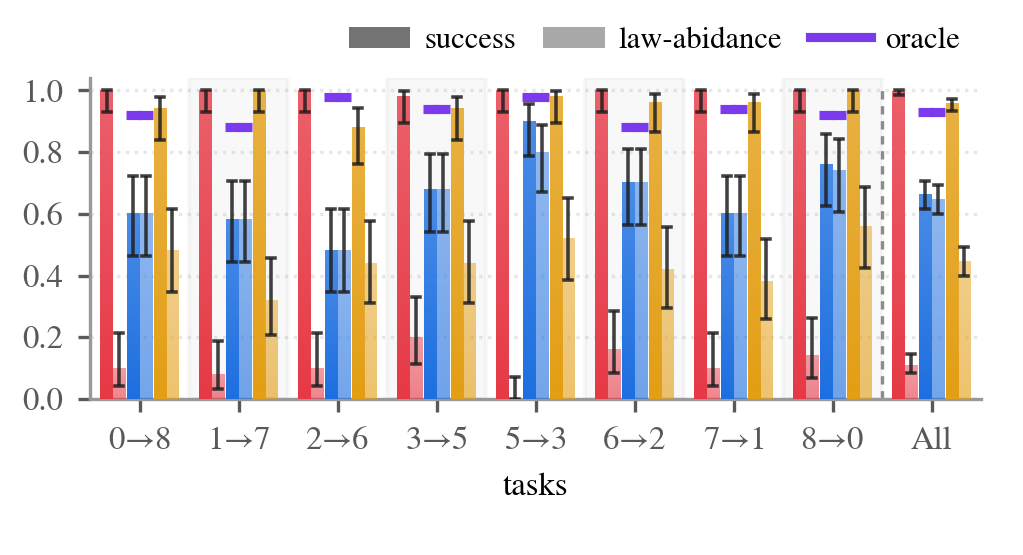}
\vspace{-24pt}
\captionof{figure}{\textbf{Baseline Experiment.} Success (solid) and law abidance (light) rates by task, with \textbf{All} pooled across tasks; bars are $95\%$ CIs and the purple line is the oracle. For the social agent abidance caps success, while the others succeed far more often than they abide. The point is that \textit{ex ante} legislation shapes behavior, and what it costs in task completion depends on how the agent treats a deontic conclusion.}\label{fig:signlawset}
\end{minipage}\hfill
\begin{minipage}[t]{0.26\linewidth}\centering
\raisebox{7.8pt}{\includegraphics[width=\linewidth]{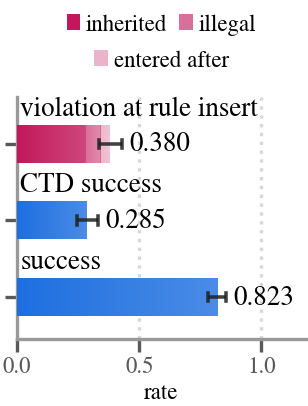}}
\vspace{-24pt}
\captionof{figure}{\textbf{Rule Insertion.} \textit{R11} inserted at $t=3$. The point is that an amendment strands the agent in violation, and the CTD duty repairs it without costing success.}\label{fig:ruleinsert}
\end{minipage}
\vspace{-5pt}
\end{figure}

\begin{figure}[!t]\centering
\begin{minipage}[c]{0.315\linewidth}\centering
  {\footnotesize (a)}\\[1pt]
  \includegraphics[height=3.25cm]{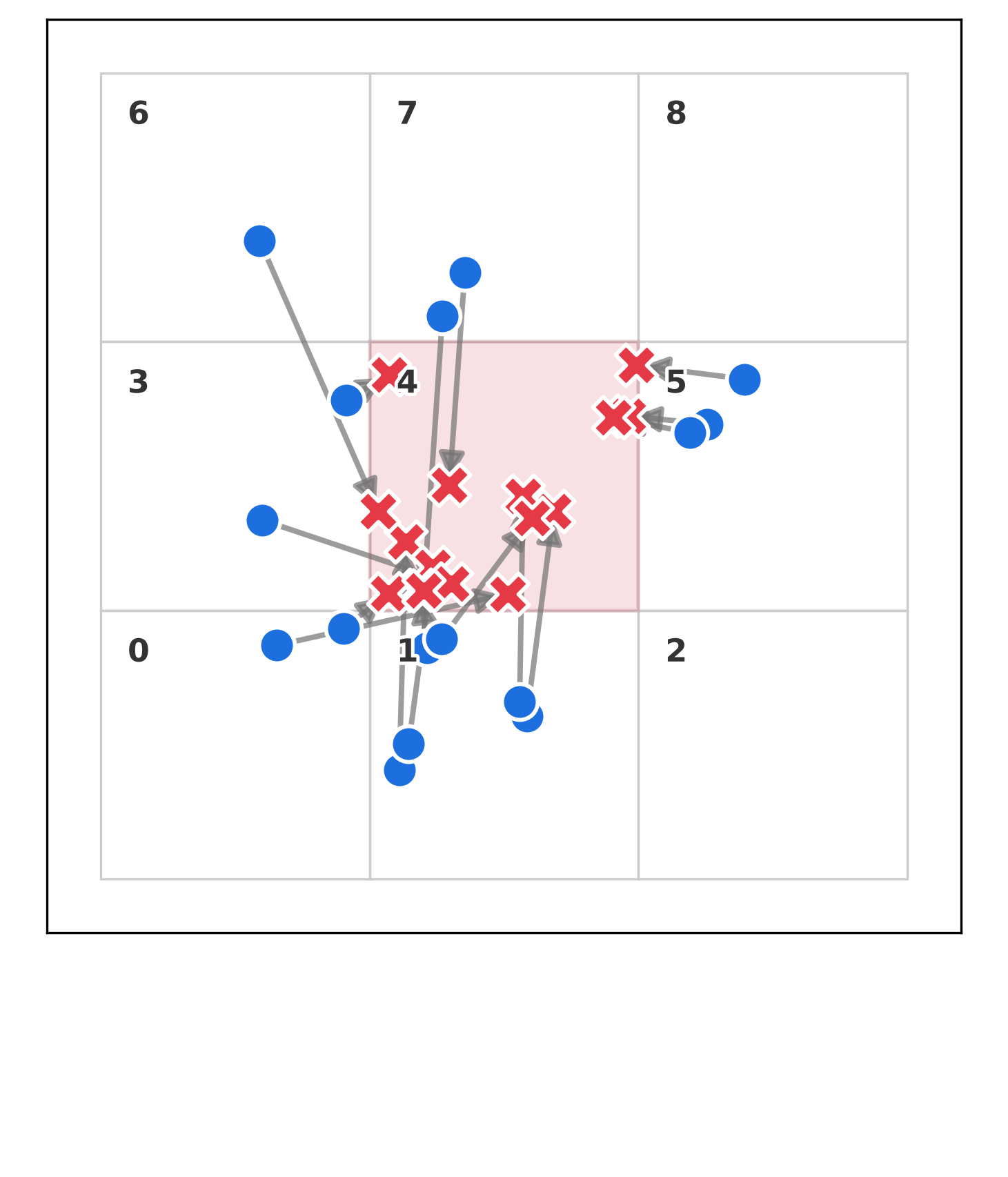}
\end{minipage}\hfill
\begin{minipage}[c]{0.20\linewidth}\centering
  \raisebox{8pt}[0pt][0pt]{\parbox{\linewidth}{\centering
  {\footnotesize (b)}\\[1pt]
  \scriptsize\setlength{\tabcolsep}{4pt}\renewcommand{\arraystretch}{1.05}
  \begin{tabular}{@{}lr@{}}
  \toprule
  \multicolumn{2}{@{}c@{}}{WM\,+\,probe error}\\
  \multicolumn{2}{@{}c@{}}{meters (cell $=0.13$)}\\
  \midrule
  Mean   & 0.031 \\
  Median & 0.019 \\
  IQR    & 0.020 \\
  p90    & 0.067 \\
  p99    & 0.190 \\
  \bottomrule
  \end{tabular}}}
\end{minipage}\hfill
\begin{minipage}[c]{0.435\linewidth}\centering
  {\footnotesize (c)}\\[1pt]
  \includegraphics[height=3.25cm]{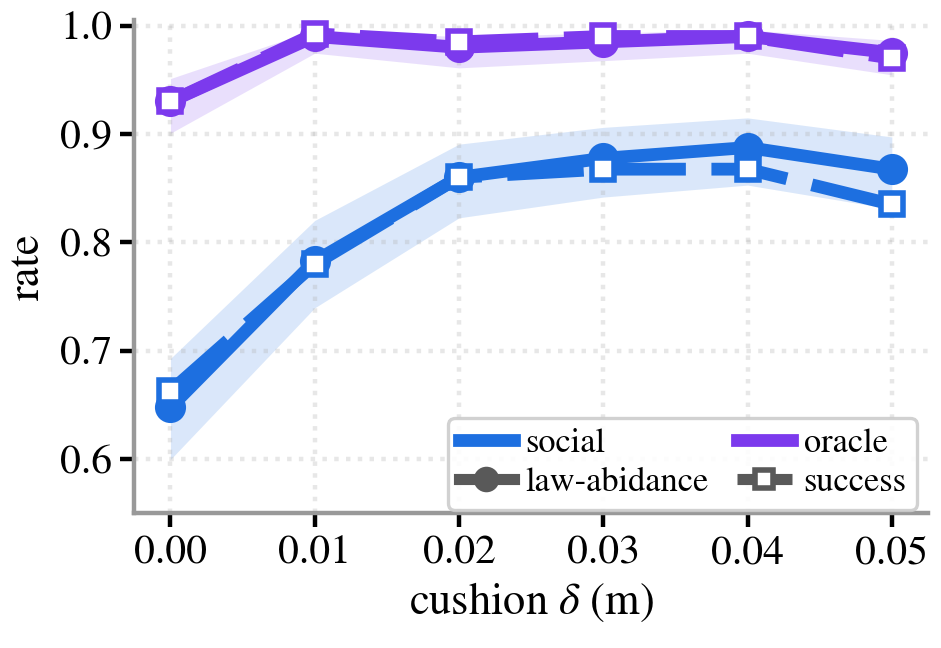}
\end{minipage}
\captionof{figure}{\textbf{The Grounding Isomorphism Gap. }(a) the $16$ largest cases (social) where predicted (\legdot) and executed (\legx) rest fall in different cells (\legarrow). (b) WM + probe error vs ground truth per \emph{executed action}, pooled over the three agents. (c) a cushion $\delta$ dilating the enforced footprint, Wilson $95\%$ bands. The point is that perception error is large enough to ground the law on the wrong cell, and modeling uncertainty narrows the gap.}\label{fig:q2}
\vspace{-5pt}
\end{figure}

Figure~\ref{fig:signlawset} shows that \textit{ex ante} legislation with DDL works, but the tradeoff between law abidance and success rates depends on agent design. The realistic and deviant agents succeed far more often than they abide, so the law does not constrain them from task completion. The social agent's success ($66.2\%$) is capped by its abidance rate ($64.7\%$), because breaking the law turns the sign red (R2, R3, R9) and prevents it from moving towards the goal. The social agent abides $5.9\times$ more often than the realistic agent, so the legal layer is what produces the compliance. Abidance is not capped there: modeling the perception uncertainty with a cushion $\delta$ (Figure~\ref{fig:q2}(c)) lifts the social agent from $64.7\%$ to $88.7\%$ at $\delta=0.04$, with success rising alongside it, so \textit{ex ante} legislation can be made substantially more effective than the baseline stack.

\paragraph{Q2: Can the Agent Adapt to Exogenous Signals?}\label{sec:Q2}

The rule sign is an exogenous signal in that the agent does not control it. The sign changes color based on the trajectory's ground truth history (R5, R7, R7b), so the facts the agent reasons over change mid episode. Table~\ref{tab:ruleeng} shows that the social agent tracks the flip. When the sign forbids the center cell (R2), it moves the cube in on $7.4\%$ of the decisions where it could have. When the sign permits it (R4), it moves in on $53.5\%$, a $7.2\times$ increase, demonstrating that the exogenous signal shapes the agent's means to its ends. 

\paragraph{Q3: Can the Agent Adapt to Endogenous Changes?}\label{sec:Q3}

Rule insertion is an endogenous change, i.e.\ one occurring within the system. Figure~\ref{fig:ruleinsert} shows that runtime rule insertion is possible, meaning legal design is updateable and does not need to be set \textit{a priori}. Its three bars are all fractions of the $400$ episodes. In $152$ of them ($0.380$) the cube was in the center when \textit{R11} fired; the figure stacks these by how it got there, and in $112$ it held a green license the reverting sign revoked underneath it, the rest having already been violating or entered on that frame. The CTD rule pushed $114$ back into a legal cell by the next time step, which is $0.285$ of all episodes and $75.0\%$ of the $152$ at risk. Finally, $329$ ($0.823$) returned to the start cell, so the fallback did not compromise task success rate.

\paragraph{Q4: The Grounding Isomorphism Gap: Does World Model error affect legislation?} \label{sec:Q4}
The main observation is that prediction error was strong enough to influence law abidance. Figure~\ref{fig:q2}(b) reports that error, measured per executed action and pooled across Social, Deviant, and Realistic runs. An average WM + probe error of 0.031m ($23\%$ cell width) is enough to convert a predicted, efficient, and legal path towards the goal into a violation. More important, however, is that the 90th percentile error rate averages 0.067m ($50\%$ cell width) and 99th percentile error reaches 0.190m ($143\%$ cell width) pooled over all committed actions, which exceeds a full cell width.

This error then results in the grounding isomorphism gap, visualized in worst case examples in Figure~\ref{fig:q2}(a). The world model predicts a legal cell (\legdot), grounding a candidate action as a legal atom that passes the DDL check, but when executed that action is illegal (\legx). The larger the tail end error, the harder it is to close the grounding isomorphism gap. The oracle gives the perception error ceiling, abiding $93\%$ of the time against the social agent's $64.7\%$ (Figure~\ref{fig:signlawset}): that $28.3\%$ gap is the price of grounding the law in learned perception, not a limit of the DDL layer.

This gap in turn results in misfiring fallback mechanisms within the law. Of R10's 129 $[\mathsf{O}] \text{exit\_cell}(4)$ conclusions for the social agent (Table~\ref{tab:ruleeng}), 61 were true violations where the cube was in the illegal cell, and 68 were probe error that caused R10's compensatory clause to misfire. The deviant agent fires 141 times; 84 were true and 57 were probe error. Specifically, the false positives carry a cost that no compliance metric records: $53\%$ of the social agent's and $40\%$ of the deviant's R10 repair obligations were levied on a cube that was never in the illegal cell, forcing the agents to pay for a violation they never committed.

Modeling this uncertainty provides a way to overcome the gap (Figure~\ref{fig:q2}(c)): a cushion $\delta$ dilating the social agent's enforced footprint makes the grounded atom absorb the prediction error instead of trusting a point estimate. Choosing an optimal cushion size is left to future work, as well as mechanisms to recover the $143\%$ tail.

\paragraph{Q5: The Ontological Isomorphism Gap: How are Reasoning Conclusions Translated to Planning Constraints?}\label{sec:Q5}
The ontological isomorphism gap is the mismatch between DDL conclusions and planning constraints. The success and abidance definitions of Section~\ref{sec:setup}, one scored on the cube's center and the other on its whole footprint, are an ad hoc engineering judgment that could easily be inverted. Table~\ref{tab:q3_ontology} re-evaluated R2 (Table~\ref{tab:full_lawset}) in the same episodes with different metrics, or interpretations of the law. The episodes are only reevaluated, not rerun.

\begin{table}[t]\centering\footnotesize
\setlength{\tabcolsep}{6pt}
\begin{tabular}{@{}lrrrrrrr@{}}
\toprule
 & & \multicolumn{2}{c}{success} & \multicolumn{2}{c}{abidance, full path} & \multicolumn{2}{c}{abidance, at rest} \\
\cmidrule(lr){3-4}\cmidrule(lr){5-6}\cmidrule(l){7-8}
Agent & $n$ & \textbf{center} & footprint & center & \textbf{footprint} & center & footprint \\
\midrule
\agreal~realistic & 400 & 0.998 & 0.998 & 0.138 & 0.113 & 0.352 & 0.130 \\
\agsoc~social  & 400 & 0.662 & 0.677 & 0.948 & 0.655 & 0.978 & 0.833 \\
\agdev~deviant & 400 & 0.958 & 0.988 & 0.557 & 0.448 & 0.680 & 0.497 \\
\bottomrule
\end{tabular}
\vspace{5pt}
\caption{\textbf{Measuring the Ontological Isomorphism Gap: }Rescoring R2 alone, not with the full lawset; bold is the implemented metric in Questions~\hyperref[sec:Q1]{1}~and~\hyperref[sec:Q4]{4}. Full path uses the line between start and end poses, at rest only the frames the planner perceives; center is the cube's center, footprint its full area. Episodes are not rerun under each metric, only reevaluated. The table shows how the metric choice requires interpretation of the law.}
\label{tab:q3_ontology}
\vspace{-10pt}
\end{table}

Because of the ontological isomorphism gap, how many metrics were possible only became apparent from building the environment and enforcing R2 in it, something a legislator drafting the rule would never see. The engineer's choice in this setup could be responsible for up to a $32.3\%$ difference in law abidance scoring for the social agent, $23.9\%$ for the realistic agent, and $23.2\%$ difference for the deviant agent.

\paragraph{Q6: Is the System Auditable and Cheap to Deploy?}\label{sec:Q6}
The stack is auditable because every deontic decision is logged with the rule that fired, the atoms grounding it, and the verdict. Comparing that ledger against ground truth separates a rule being in force from its being obeyed, attributing failures to perception or to reasoning. In Table~\ref{tab:ruleeng}, R2 is enforced $91.1\%$ of the time for the social agent and $87.3\%$ for the deviant. R10's misfires are due to perception error, noted as false positives (\hyperref[sec:Q4]{Question 4}).

\begin{table}[th]\centering\footnotesize
\setlength{\tabcolsep}{2pt}
\begin{tabular}{@{}lrrrrrr@{}}
\toprule
 & \multicolumn{2}{c}{Social} & \multicolumn{2}{c}{Deviant} & \multicolumn{2}{c}{Realistic} \\
\cmidrule(lr){2-3}\cmidrule(lr){4-5}\cmidrule(l){6-7}
Rule & Engaged & Enforced & Engaged & Enforced & Engaged & Enforced \\
\midrule
R2 \texttt{no\_center\_cell}         & 841  & 766 ($91.1\%$)  & 769 & 671 ($87.3\%$) & 556 & 396 ($71.2\%$) \\
R5 \texttt{yellow\_sign}             & 824  & 824 ($100.0\%$) & 741 & 732 ($98.8\%$) & 492 & 451 ($91.7\%$) \\
R9 \texttt{conflicting\_permissions} & 1311 & 1311 ($100.0\%$)& 568 & 215 ($37.9\%$) & 523 & 0 ($0.0\%$)    \\
R10 \texttt{contrary\_to\_duty}      & 129  & 113 ($87.6\%$)  & 141 & 133 ($94.3\%$) & 203 & 59 ($29.1\%$)  \\
\quad \emph{of which false positive} & 68   & ---             & 57  & ---            & 58  & ---            \\
\midrule
Rule & At risk & Entered & At risk & Entered & At risk & Entered \\
\midrule
R2 \texttt{no\_center\_cell}         & 766  & 57 ($7.4\%$)    & 671 & 137 ($20.4\%$) & 396 & 288 ($72.7\%$) \\
R4 \texttt{green\_sign}              & 417  & 223 ($53.5\%$)  & 338 & 152 ($45.0\%$) & 101 & 36 ($35.6\%$)  \\
\bottomrule
\end{tabular}
\vspace*{6pt}
\caption{\textbf{Pipeline audit}: \emph{Engaged}: the rule's conclusion was in force in the runtime ledger; \emph{Enforced}: it also
held in ground truth. The lower block reports uptake: \emph{At risk} is decisions taken with the cube outside the center, so entry was possible; \emph{Entered} is how many of those decisions put it in the center at the next step. All counts are over steps taken while the episode was still running; frames logged after the goal is reached are excluded. }
\label{tab:ruleeng}
\vspace{-10pt}
\end{table}

Reaching a verdict is cheap: symbolic reasoning costs $16.6$ms per deontic decision, orders of magnitude less than the $12.5$s the planner spends on each executed action.

\vspace{-15pt}
\section{Related Work}\label{sec:relatedwork} 
\vspace{-5pt}
This work is generally motivated by the AI Safety, Alignment, and Ethics (AI SAE) agenda that grounds alignment in evolutionary biology \cite{waldner2025aisae}.

Grounding in the legal AI literature, deontic logic has been used to generate normative penalties for RL policies \cite{neufeld2024bolts}, to ensure compliant goals \cite{neufeld2022ethical}, and extended to DDL theories to supervise agent actions \cite{neufeld2021normative}. Previous work has also focused on applying the law \textit{ex ante} to learning agents \cite{riveret2013exante}. DDL has been used for legislating autonomous vehicles in simulation \cite{bhuiyan2024traffic}. An implementation of DDL in Answer Set Programming (ASP) \cite{Governatori2024ASPDDL} has been used to do planning \cite{sartor2025plans}. Norms have also been amended at runtime, but symbolically and without perception \cite{olson2026norms}. This work applies deontic logic to sample based motion planners in simulation for the first time.

Theoretically, the legal planning stack is grounded in the lineage of the ethical governor \cite{arkin2009ethical}. The \emph{ex-ante} enforcement is closest in spirit to safety shielding \cite{alshiekh2018safe,carr2023shielding}, robot motion planning with temporal logic specifications \cite{ZHANG2020105591,luo2021abstraction,kantaros2020stylus,vasile2020reactive,kudalkar2026sampling}, and minimum violation temporal logic planning  \cite{MVRRT,cai2023learning,tumova2016least,vasile2017sampling,buyukkocak2025resilient}. However, defeasible deontic logic is used here because it explicitly represents exceptions, duties and violations \cite{Nute1997-NUTDDL-3, ddl}. Instead of filtering the search space pre generation, nodes are adjudicated post generation because legislation depends on observations and predictions, not the state.

Similar robot safety work uses latent space planning with Hamilton-Jacobi reachability on top of DreamerV3 \cite{hafner2025mastering} to compute safety preserving filters and control in a manipulation task \cite{NakamuraK-RSS-25}. There is precedent for using probes to ground neural networks' internal representations in propositions (e.g. Safe/Unsafe) for runtime alignment \cite{krishna2026disentangled, NEURIPS2023_81b83900, ICLR2025_3132d040}. Neuro-symbolic predicate world models \cite{liang2025visualpredicator, liang2026exopredicator, athalye2026pixels} learn their symbolic vocabulary to make planning tractable; ours is fixed by the legal source text. Embedding Temporal Logic \cite{kapoor2026etl} goes further and replaces propositions with distances in a learned embedding space, which shifts the grounding problem rather than removing it. Keeping the vocabulary propositional is what lets every verdict here be traced back to the atoms that produced it.

\vspace{-15pt}
\section{Discussion and Future Work}\label{sec:future}
\vspace{-5pt}
The grounding isomorphism gap is an engineering problem that undermines the robustness of robot law at deployment. The ontological isomorphism gap falls in the seam between two fields: enforcing even a seemingly brightline rule requires judicial reasoning more technical than what legislators are expected to consider, and more jurisprudential than what engineers are typically held responsible for. 

The cushion sweep (Figure~\ref{fig:q2}(c)) closes part of the grounding gap, lifting abidance from $64.7\%$ to $88.7\%$, but not the $143\%$ tail; reaching it needs conformal prediction, which may be used to model the \textit{types} of error in the system, e.g. tail end prediction error and biases towards specific states. For validation, deployment of the legal system into physical robotics instead of simulation will be critical for detecting failure modes in the real world; a \textit{Sim to Real Isomorphism gap} likely exists as well. The ontological gap needs each rule shipped with the operationalization it was drafted against, making the engineer's choice auditable.

\vspace{-15pt}
\section{Conclusion}\label{sec:conc}
\vspace{-5pt}
This research implemented a system of robot governance idiosyncratic to world-model based planning systems for robotics, leveraging the latent prediction space to perform \textit{ex ante} governance. This work establishes that deployment-ready governance faces the grounding isomorphism gap and the ontological isomorphism gap: an ostensibly brightline rule can vary widely in abidance depending on the metric. By closing these gaps, robot law can be deployed as legislatures intend it to align robot behavior with society's norms.

\clearpage

\bibliographystyle{vancouver}
\bibliography{references}
\end{document}